\documentclass[10pt,twocolumn,letterpaper]{article}

\usepackage{cvpr}              
\usepackage{booktabs}
\usepackage{adjustbox}
\usepackage[table]{xcolor}
\usepackage{float}
\usepackage{xcolor}
\usepackage{multirow}
\providecolor{TableGroup}{HTML}{F2F3F5}
\providecolor{TableText}{HTML}{202B3A}
\definecolor{TableOurs}{HTML}{E7F0FB}

\definecolor{cvprblue}{rgb}{0.21,0.49,0.74}
\usepackage[pagebackref,breaklinks,colorlinks,allcolors=cvprblue]{hyperref}

\def\confName{CVPR}
\def\confYear{2027}

\title{Synergizing Evidence Retrieval and Utilization for Self-Improving Multimodal Scientific Document Reasoning}

\author{
    {\large
    Frank Nie$^{1,*}$ \quad
    Ethan B. Liu$^{1,*,\dagger}$ \quad
    Yuan Zhu$^{1,*}$ \quad
    Jindong Han$^{1,\dagger}$
    } \\
    \\
    $^{1}$Shandong University, China \\
    {\small $^{*}$Equal contribution. \quad
    $^{\dagger}$Corresponding author.}
}

\begin{document}
\maketitle

\begingroup
\renewcommand{\thefootnote}{\fnsymbol{footnote}}
\footnotetext[1]{Equal contribution.}
\footnotetext[2]{Corresponding author.}
\endgroup

\begin{abstract}
Scientific document reasoning requires an agent to acquire and integrate evidence distributed across text, figures, and tables. Retrieval can identify relevant sources while missing complementary support, and answer-level feedback alone does not reveal whether an error stems from evidence acquisition or evidence use. We present an agentic Retriever--Perceiver--Reasoner framework for multimodal scientific document reasoning. The Retriever sequentially selects source-aligned evidence under a context budget. Given the selected evidence and question, the Perceiver constructs a question-conditioned latent ontology, an implicit organization of multimodal evidence in a shared continuous space. The Reasoner combines this latent guidance with the original sources to produce grounded answers. We train the three components through diagnostic-guided co-evolution: alternating updates optimize them with a shared answering reward, while Retrieval Potential Feedback (RPF) and Latent Efficacy Feedback (LEF) identify questions that need better retrieval or latent guidance and direct targeted resampling. Across four benchmarks and seven evaluation settings, our framework obtains average scores of 69.9 and 75.0 with 7B and 8B Reasoner backbones, exceeding the strongest baseline by 4.0 and 9.1 points, respectively.
\end{abstract}

\section{Introduction}
\label{sec:am_introduction}

\begin{figure}[t]
\centering
\includegraphics[width=0.95\columnwidth]{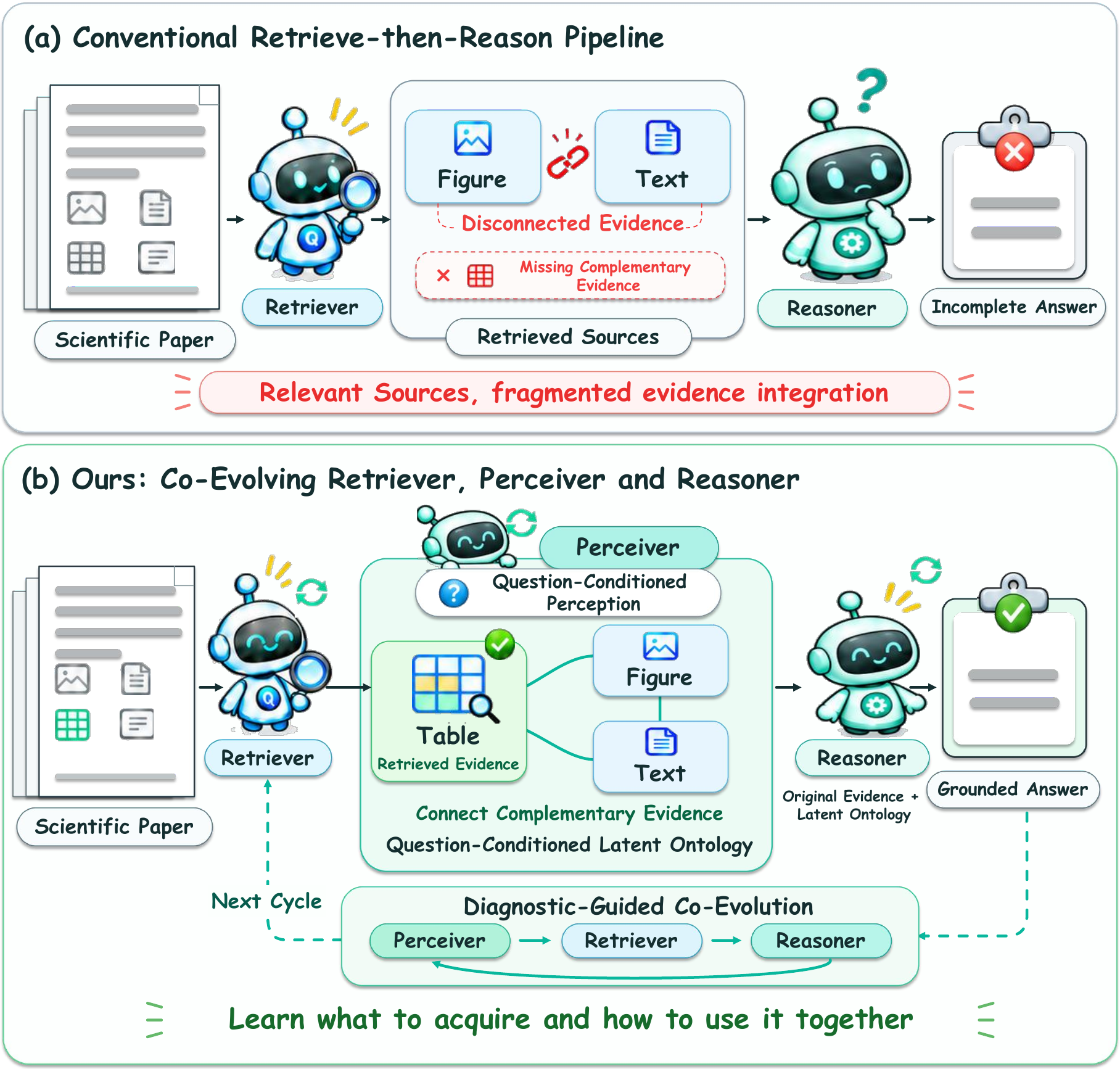}
\caption{Motivation and overview of our agentic framework. (a) Conventional retrieve-then-reason pipelines may retrieve relevant sources without integrating complementary multimodal evidence. (b) Our framework couples relation-aware retrieval with question-conditioned latent ontology construction, allowing the Reasoner to combine latent guidance with original sources. Diagnostic-guided co-evolution further coordinates evidence acquisition and utilization through a shared answering reward and paired diagnostics.}
\label{fig:am_motivation}
\end{figure}

Scientific document reasoning enables researchers to interpret prior findings and identify promising directions for future research~\citep{lo2023semantic}. Answering a question about a scientific paper requires identifying its claims, the evidence supporting them, and the conditions under which they hold~\citep{chen2026scimdr,dasigi2021dataset}. Because this information is distributed across text, figures, and tables, reliable answers require both acquiring sufficient evidence and integrating complementary information across modalities~\citep{pramanick2024spiqa,masry2022chartqa}. Although multimodal large language models provide a powerful foundation for this task, access to the full document does not ensure that the necessary evidence will be located or used effectively~\citep{ma2024mmlongbench,yan2026docseeker}.


Existing approaches address evidence acquisition and utilization from complementary directions. Retrieval and evidence localization methods~\citep{yu2025visrag,cho2024m3docrag,yan2026docseeker} identify relevant document content, whereas evidence-guided and agentic reasoning methods~\citep{sun2026mitigating,wang2025vidorag} gather or organize evidence for answering. The two capabilities, however, remain loosely coupled: a relevant retrieved set may still lack complementary support, while available text, figures, and tables may not be brought together into a coherent answer. 
As Figure~\ref{fig:am_motivation}(a) illustrates, a one-way retrieve-then-reason pipeline cannot tell whether an incomplete answer reflects missing evidence or a failure to use what has already been retrieved. Better interpretation cannot recover evidence that was never retrieved, while the value of a source depends on how effectively the current model can interpret it. This reciprocal dependence motivates learning evidence acquisition and utilization together.


To bring these capabilities together, we propose an agentic framework built around a Retriever-Perceiver-Reasoner architecture, illustrated in Figure~\ref{fig:am_motivation}(b). The retriever sequentially selects passages, figures, and tables while preserving their original document locations. Each selection is conditioned on the question, the evidence acquired so far, and the remaining retrieval budget. Given the selected evidence, the perceiver performs on-demand multimodal latent perception to construct a question-conditioned latent ontology, i.e., an implicit organization of relationships among textual and visual evidence. The reasoner then uses the latent guidance alongside the original retrieved evidence to generate a grounded answer, combining cross-modal organization with direct access to source details.


We train the three components through a co-evolutionary agentic training strategy. Following supervised initialization, we alternately update the perceiver, retriever, and reasoner using a shared answer-quality reward. This procedure allows improvements in perception and reasoning to reshape which evidence is useful to retrieve, while updates to retrieval change the evidence available for latent organization and answer generation. Since answer-level reward provides limited supervision about which capability constrains performance, we further introduce two diagnostic signals. Retrieval Potential Feedback (RPF) compares answer quality obtained from annotated and retrieved evidence under the same perceiver and reasoner. A positive gap in favor of annotated evidence indicates that improving retrieval could improve answer quality. Latent Efficacy Feedback (LEF) compares answer quality without and with latent guidance under the same retrieved evidence. A positive gap in favor of omitting latent guidance indicates that the current guidance, or how it is used, may reduce answer quality. We use RPF and LEF to direct targeted resampling toward the retriever and the perceiver-reasoner pair, respectively. Together, alternating component updates and diagnostic-guided resampling coordinate the self-improvement of evidence acquisition and utilization.


Our contributions are:
\begin{itemize}
\item We introduce a retriever-perceiver-reasoner agentic framework for multimodal scientific document reasoning that coordinates adaptive evidence retrieval, on-demand latent perception, and grounded reasoning over the original retrieved evidence.
\item We develop a co-evolutionary agentic training strategy that alternates updates of the three components and incorporates dual diagnostic feedback to prioritize learning from instances with retrieval limitations or ineffective latent guidance.
\item We evaluate our framework on four benchmarks across seven settings. With 7B and 8B reasoner backbones, it achieves average scores of 69.9 and 75.0, outperforming the strongest baseline by 4.0 and 9.1 points, respectively.
\end{itemize}
\section{Related Work}
\label{sec:am_related_work}

\paragraph{Scientific Multimodal Document Reasoning.}
Scientific document understanding requires connecting findings expressed through text, figures, and tables. ChartQA~\citep{masry2022chartqa} evaluates visual and numerical chart reasoning, while CharXiv~\citep{wang2024charxiv} targets chart understanding and reasoning in scientific papers. Multimodal ArXiv~\citep{li2024multimodal} and MMSci~\citep{li2024mmsci} extend the evaluation of scientific figure understanding across disciplines. SPIQA~\citep{pramanick2024spiqa} places figure and table questions within the context of scientific papers, while M3SciQA~\citep{li2024m3sciqa} extends evaluation to reasoning across scientific papers and their cited documents. SciMDR~\citep{chen2026scimdr} provides evidence-grounded reasoning data for full-document question answering. Together, these benchmarks highlight the need to locate question-relevant sources across documents and integrate their textual and visual information to produce grounded answers.

\paragraph{Evidence Grounding and Utilization.}
Existing approaches address evidence acquisition and utilization from complementary perspectives. MIRAGE~\citep{wu2025visual}, VisRAG~\citep{yu2025visrag}, and M3DocRAG~\citep{cho2024m3docrag} explore multimodal evidence retrieval, while DocSeeker~\citep{yan2026docseeker}, VisRAG2.0~\citep{sun2026mitigating}, and ViDoRAG~\citep{wang2025vidorag} further investigate evidence localization, grounding, and iterative reasoning. Beyond retrieval and inference-time evidence handling, CLaRa~\citep{he2025clara} jointly optimizes retrieval and generation, while PRCO~\citep{miao2026seeing} co-evolves perception and reasoning. Despite these advances, acquiring relevant sources does not guarantee effective integration of complementary multimodal evidence, and answer-level feedback alone provides limited insight into whether failures arise from evidence acquisition or utilization. Our framework addresses these challenges by coupling adaptive evidence retrieval with question-conditioned latent ontology construction and introducing paired diagnostics to guide component-specific learning during co-evolution.

\begin{figure*}[t]
    \centering
    \includegraphics[width=0.82\textwidth]{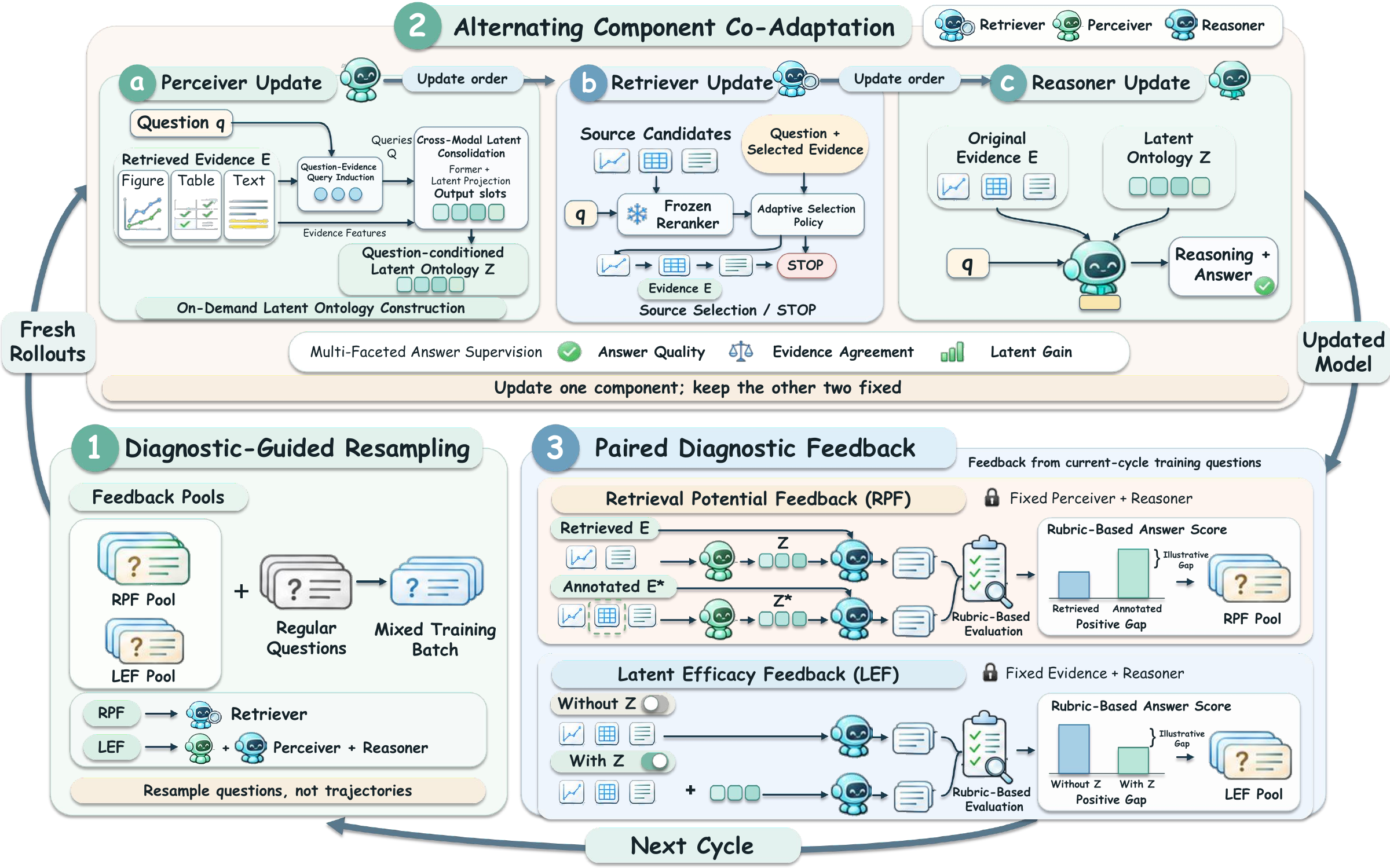}
\caption{Overview of diagnostic-guided co-evolution. (1) Diagnostic-guided resampling combines regular questions with RPF- or LEF-selected questions for component-specific training. (2) Alternating component co-adaptation sequentially updates the Perceiver, Retriever, and Reasoner under a shared answering reward, keeping the other two components fixed at each update. (3) Paired diagnostic feedback uses RPF to compare responses based on retrieved and annotated evidence, and LEF to compare responses with and without latent guidance. Positive score gaps determine the feedback pools for the next training cycle.}
    \label{fig:framework}
\end{figure*}
\section{Method}
\label{sec:am_method}

\subsection{Agentic Task Formulation}
\label{sec:am_problem}

We study agentic question answering over long multimodal scientific documents. Given a complete document $D=(X,V)$ and a question $q$, the agent generates an evidence-grounded response $y\sim p_{\Omega}(\cdot\mid D,q)$, comprising a reasoning trace and a final answer. Here, $X$ denotes the document text, $V$ contains its figures and tables with captions, and $\Omega$ denotes the trainable parameters.

The agent operates through three stages, implemented by a Retriever, a Perceiver, and a Reasoner. Given $D$, it first constructs a source-aligned candidate pool $\mathcal C(D)$ of passages, figures, and tables. At step $t$, the agent state $s_t$ encodes the current phase $\xi_t\in\{\mathrm{s},\mathrm{p},\mathrm{r}\}$, the question, selected sources, candidate features, and the latent representation when available. The phase-dependent action space is $\mathcal A(s_t)=\mathcal A_{\xi_t}(s_t)$, defined as
\begingroup
\small
\begin{equation}
    \begin{aligned}
        \mathcal A_{\mathrm{s}}(s_t)
            &=\{a_e:e\in\mathcal C_t\}\cup\{a_{\mathrm{stop}}\},\\
        \mathcal A_{\mathrm{p}}(s_t)
            &=\{a_{\mathrm{p}}\},\quad
        \mathcal A_{\mathrm{r}}(s_t)
            =\{a_{\mathrm{r}}(y):y\in\mathcal Y\}.
    \end{aligned}
    \label{eq:am_agent_actions}
\end{equation}
\endgroup
Here, $\mathcal C_t\subseteq\mathcal C(D)$ denotes the eligible candidates, and $\mathcal Y$ is the response space. Starting from $S_0=\varnothing$, the Retriever uses its policy $\pi_{\phi}(\cdot\mid s_t)$ to sequentially select sources, updating $S_{t+1}=S_t\cup\{e\}$, or terminates retrieval by selecting $a_{\mathrm{stop}}$. Once retrieval terminates at step $T$, the selected sources are packed into evidence $E$, and the Perceiver constructs a question-conditioned latent ontology $Z$. The Reasoner then generates a response using both the original evidence and the latent ontology:
\begingroup
\small
\begin{equation}
    E=\operatorname{Pack}(S_T),\quad
    Z=C_{\psi}(E,q),\quad
    y\sim p_{\theta}(\cdot\mid E,q,Z).
    \label{eq:am_pipeline}
\end{equation}
\endgroup
The latent ontology provides complementary guidance for multimodal evidence integration, while the original sources remain directly accessible to the Reasoner. The parameter blocks $\phi$, $\psi$, and $\theta$ govern the Retriever, Perceiver, and Reasoner, respectively, with $\Omega=\{\phi,\psi,\theta\}$. Figure~\ref{fig:framework} illustrates the overall architecture and co-evolutionary training procedure.

\subsection{Synergistic Evidence Acquisition and Perception}
\label{sec:am_perception}

Effective multimodal reasoning requires not only acquiring relevant evidence but also integrating complementary information across modalities. These two capabilities are interdependent: retrieved sources may remain poorly connected, while improved interpretation cannot recover missing evidence. We therefore couple adaptive source retrieval with on-demand latent perception.

\subsubsection{Relation-Aware Adaptive Retrieval}
\label{sec:am_retrieval}

We introduce a Retriever that sequentially selects passages, figures, and tables from the complete document to support subsequent perception and reasoning. Given a document $D$, a fixed rule-based procedure first constructs a source-aligned candidate pool $\mathcal C(D)$ independently of the question and gold annotations. Each candidate retains its source identity and, for textual content, its original source interval. The same candidate construction and packing procedures are applied during training and inference (Appendix~\ref{sec:am_appendix_evidence}).

The Retriever selects evidence conditioned on the question, previously acquired sources, and remaining retrieval budget, subject to limits of $K_{\max}$ source units and $B_E$ evidence tokens. A frozen multimodal reranker provides question--candidate representations and relevance scores, which are fused with candidate, selection, and budget features. To capture document structure and cross-candidate dependencies, we employ two complementary mechanisms: explicit structural modeling and global candidate interactions. Specifically, we construct a source-relation graph with candidates as nodes, connecting overlapping textual blocks within the same section and linking text to referenced figures and tables, with self-loops preserving candidate-specific information. A message-passing layer propagates structural information over the graph, after which a candidate-level Transformer captures global interactions among candidates. Finally, candidate-selection and stopping heads produce action scores $u_{\phi}(a,s_t)$, defining the retrieval policy as
\begingroup
\small
\begin{equation}
\pi_{\phi}(a\mid s_t)=
\frac{\exp(u_{\phi}(a,s_t)/\tau_r)}
{\sum_{b\in\mathcal A_{\mathrm{s}}(s_t)}
\exp(u_{\phi}(b,s_t)/\tau_r)},
\label{eq:am_retrieval_policy}
\end{equation}
\endgroup
where $\tau_r>0$ is the sampling temperature.

At each step, the eligible set $\mathcal C_t$ excludes previously selected, incompatible, and over-budget candidates. Retrieval terminates when the policy selects $a_{\mathrm{stop}}$, the source limit is reached, or no eligible candidates remain. The selected sources are then packed into evidence $E$, merging overlapping textual intervals while preserving the original visual content for subsequent perception and reasoning.

\subsubsection{Question-Conditioned Latent Ontology}
\label{sec:am_latent}

To realize this complementary perception, we introduce a Perceiver that organizes retrieved multimodal evidence into a shared latent space. We term this implicit, question-conditioned organization a \emph{latent ontology}. Given the retrieved evidence $E$ and question $q$, the Perceiver constructs $Z=C_{\psi}(E,q)$, learning to organize question-relevant multimodal evidence through supervision from the Reasoner's answering objective.

\textbf{Perceiver: Latent Ontology Learning.}
The Perceiver comprises two components: a QueryBuilder for evidence-aware query generation and a Latent Evidence Consolidator for cross-modal integration.

\textit{(i) Question-Evidence Query Induction.}
Effective perception must account for both the question and the retrieved evidence. Let $H_E$ and $H_q$ denote their respective features extracted by the frozen multimodal backbone. The QueryBuilder uses $N_q$ learnable query slots to attend jointly to both representations and generate evidence-aware queries $Q$, guiding subsequent consolidation toward information relevant to the question and available in retrieved sources.

\textit{(ii) Cross-Modal Latent Consolidation.}
Guided by $Q$, the Latent Evidence Consolidator organizes multimodal evidence directly in hidden-state space, without an intermediate textual summary. Specifically, we concatenate the evidence features $H_E$, query tokens $Q$, and $M$ learnable output slots $U$, and process them with a backbone-specific Former that reuses the corresponding pretrained language-model decoder with a dedicated Perceiver LoRA adapter. We extract the output-slot hidden states and project them into the Reasoner's embedding space to obtain the latent ontology:
\begingroup
\small
\begin{equation}
\begin{gathered}
Q = \operatorname{QB}_{\psi}(H_E,H_q),\quad
H_U = \operatorname{Former}_{\psi}([H_E;Q;U])\big|_U,\\
Z = H_U W_Z+b_Z\in\mathbb{R}^{M\times d}.
\end{gathered}
\label{eq:am_latent}
\end{equation}
\endgroup
Here, $[\cdot;\cdot]$ denotes sequence concatenation, $\big|_U$ selects the output-slot hidden states, and $d$ is the Reasoner's embedding dimension.

\textbf{Reasoner: Complementary Evidence Integration.}
Although the latent ontology organizes question-relevant multimodal information, its fixed-length representation may not preserve all source-level details. The Reasoner therefore conditions generation on the question $q$, original evidence $E$, and latent ontology $Z$. While $Z$ provides compact cross-modal guidance, $E$ retains the original textual and visual details needed for grounded reasoning. Specifically, we append $Z$ to the native multimodal embeddings of the evidence--question prompt and jointly prefill the resulting sequence before autoregressively generating the reasoning trace and final answer. This design allows latent guidance to complement, rather than replace, the original evidence.

\subsection{Diagnostic-Guided Co-Evolutionary Training}

\paragraph{Reasoning-Aligned Perceiver Initialization.}
The Perceiver is initialized to organize multimodal evidence in a form that facilitates downstream reasoning. We use 7,000 examples from the SciMDR training set~\citep{chen2026scimdr}, containing questions, annotated evidence, and reference responses. To isolate perception learning from retrieval errors, we bypass the Retriever and directly provide annotated evidence $E^*$. The Perceiver constructs $Z=C_{\psi}(E^*,q)$, which is jointly processed with the original evidence by a frozen Reasoner. Using teacher forcing, we compute the mean per-token negative log-likelihood $\ell_Z$ of the reference response $y^*$, including its reasoning trace and final answer. The initialization objective is
$\mathcal{L}_{\mathrm{I}}=\ell_Z.$
Only the Perceiver parameters $\psi$ are updated, with gradients propagated through the frozen Reasoner to optimize the latent ontology.

\paragraph{Evidence-Aligned Retriever Initialization.}
The Perceiver initialization provides a reasoning-aligned representation, but the Retriever must additionally learn to acquire supporting evidence from the complete document. We therefore initialize the Retriever using the same 7,000 SciMDR training examples. Specifically, we align the annotated evidence $E^*$ with the automatically constructed candidate pool $\mathcal C(D)$ from Section~\ref{sec:am_retrieval}, establishing source-level supervision for retrieval.

Rather than prescribing a fixed selection order, we evaluate each action by the quality of the resulting evidence set. At state $s_t$, let $S_t\oplus a$ denote the selected set after taking action $a$, with the STOP action leaving $S_t$ unchanged. The target action score is defined as
$v_{t,a}=F_{\mathrm{ret}}(S_t\oplus a,E^*),$
where $F_{\mathrm{ret}}$ equally averages source-aligned text and visual F1 over active modalities, balancing annotated-evidence coverage against the inclusion of unannotated content. Since multiple actions may yield similarly effective evidence sets, we adopt set-valued supervision rather than enforcing a unique selection order. Specifically, for a tolerance $\delta\geq 0$, we define the near-optimal action set as
\begingroup
\small
\begin{equation}
    \mathcal A^{+}(s_t)=
    \left\{a\in\mathcal A_{\mathrm{s}}(s_t):
    v_{t,a}\geq
    \max_{b\in\mathcal A_{\mathrm{s}}(s_t)}v_{t,b}-\delta
    \right\}.
    \label{eq:am_near_opt_actions}
\end{equation}
\endgroup
The initialization objective combines action-score regression with probability-mass supervision over this set:
\begingroup
\small
\begin{equation}
\begin{split}
\mathcal{L}_{\mathrm{II}}(s_t)
={}&\frac{1}{|\mathcal A_{\mathrm{s}}(s_t)|}
\sum_{a\in\mathcal A_{\mathrm{s}}(s_t)}
\bigl(u_{\phi}(a,s_t)-v_{t,a}\bigr)^2\\
&-\lambda_a\log
\sum_{a\in\mathcal A^{+}(s_t)}
\pi_{\phi}(a\mid s_t),
\end{split}
\label{eq:am_stage2_loss}
\end{equation}
\endgroup
where $\lambda_a\geq 0$ balances score calibration and action selection. Training covers both annotation-guided selection paths and policy-visited partial selections. Only the Retriever parameters $\phi$ are updated, while the reranker and other components remain frozen. Further details of evidence alignment and scoring are provided in Appendix~\ref{sec:am_appendix_evidence}.

\paragraph{Diagnostic-Guided Co-Evolution.}
\label{sec:am_stage3}

Following supervised initialization, the three components must co-adapt: retrieval determines the evidence available for perception and reasoning, while downstream answering provides feedback for improving retrieval. We therefore introduce a diagnostic-guided co-evolution strategy that alternates updates of the three components under a shared answering reward, with paired diagnostics directing subsequent learning.

\textbf{Alternating Component Co-Adaptation.}
Starting from the initialized Retriever and Perceiver, we initialize a separate Reasoner adapter with parameters $\theta$ and sequentially update the three components:
\begingroup
\small
\begin{equation}
    \Omega^{k+1}
    =\bigl(\mathcal U_{\theta}\circ
           \mathcal U_{\phi}\circ
           \mathcal U_{\psi}\bigr)(\Omega^k),
    \label{eq:am_cycle}
\end{equation}
\endgroup
where $\mathcal U_b$ updates only parameter block $b$, keeping the other components fixed. Following the right-to-left update order, the Perceiver first adapts its latent ontology to the currently retrieved evidence while the Reasoner remains fixed. The Retriever then updates its selection policy using answering feedback from the updated Perceiver and fixed Reasoner. Finally, the Reasoner adapts to the updated retrieval results and latent guidance. Each update uses fresh rollouts generated by the current model, and the resulting parameters are carried forward to the next update.

\textbf{Multi-Faceted Answer Supervision.}
To jointly optimize evidence acquisition, latent perception, and reasoning, we introduce a shared reward that evaluates both answer quality and the contributions of retrieved evidence and latent guidance. For open-ended scientific question answering, we construct reference-grounded rubrics following prior work~\citep{gunjal2026rubrics,yu2026reinforcement}. Given a question $q$ and reference answer $a^*$, a language-model compiler decomposes the answer into independently assessable claims with importance weights and partial-credit anchors. A judge evaluates a generated response $y$ as
\begingroup
\small
\begin{equation}
    S(y\mid q)=
    \frac{\sum_{j=1}^{J_q}w_j g_j(y)}
         {\sum_{j=1}^{J_q}w_j},
    \label{eq:am_rubric_score}
\end{equation}
\endgroup
where $J_q$ is the number of claims, $w_j>0$ denotes the importance weight, and $g_j(y)\in[0,1]$ measures the satisfaction of claim $j$. Further details of rubric construction are provided in Appendix~\ref{sec:am_appendix_rubrics}.

For each question, we sample $G$ retrieval trajectories, yielding evidence sets $E_i$ and corresponding latent ontologies $Z_i=C_{\psi}(E_i,q)$. For each trajectory, the Reasoner generates paired responses with and without $Z_i$, keeping the evidence, question, and decoding settings identical. Let $S_i^Z$ and $S_i^0$ denote their respective rubric scores. We define the shared reward as
\begingroup
\small
\begin{equation}
    R_i=S_i^Z
    +\lambda_{\mathrm{ret}}F_{\mathrm{ret}}(E_i,E^{*})
    +\lambda_{\mathrm{gain}}(S_i^Z-S_i^0),
    \label{eq:am_reward}
\end{equation}
\endgroup
where the three terms measure answer quality, agreement with annotated evidence, and the contribution of latent guidance, respectively. The reward weights remain fixed across components and training cycles. We optimize the Perceiver, Retriever, and Reasoner in turn using GRPO~\citep{shao2024deepseekmath}, updating only the active component while keeping the other two fixed. Annotated evidence and reference-grounded rubrics are used exclusively for reward computation.

\begin{table*}[t]
    \centering
   \caption{Comparison with general-purpose MLLMs, reasoning-oriented MLLMs, document-specialized models, and task-adapted backbones. Avg.\ is the unweighted mean over seven settings. \textbf{Bold} and \underline{underline} denote the best and second-best results in each column.}
    \label{tab:main_results}
    \small
    \setlength{\tabcolsep}{3pt}
    \renewcommand{\arraystretch}{1.16}
    \scalebox{0.75}{%
    \begin{tabular}{l c *{8}{c}}
        \toprule
        \textbf{Method} & \textbf{Backbone Size} & \textbf{ChartQA}
        & \textbf{CharXiv-D} & \textbf{CharXiv-R} & \textbf{SPIQA-A}
        & \textbf{SPIQA-B} & \textbf{SPIQA-C} & \textbf{SciMDR-Eval}
        & \textbf{Avg.} \\
        \midrule
        \rowcolor{TableGroup}
        \multicolumn{10}{l}{\textcolor{TableText}{\textbf{General-purpose MLLMs}}} \\
        Qwen2.5-VL-7B & 7B & 86.4 & 65.0 & 37.7 & 66.4 & 56.6 & 48.9 & 19.8 & 54.4 \\
        Qwen3-VL-8B-Instruct & 8B & 87.4 & 74.2 & 40.1 & 73.2 & 64.0 & 62.3 & 34.2 & 62.2 \\
        Keye-VL-1.5-8B & 8B & 85.8 & 65.8 & 39.4 & 59.6 & 56.1 & 49.3 & 65.2 & 60.2 \\
        LLaVA-OneVision-1.5-8B & 8B & 85.9 & 66.3 & 32.9 & 66.0 & 62.7 & 51.1 & 15.5 & 54.3 \\
        InternVL3-8B & 8B & 86.2 & 66.7 & 34.6 & 59.6 & 46.9 & 40.8 & 16.8 & 50.2 \\
        InternVL3.5-8B & 8B & 85.4 & 70.6 & 40.5 & 67.1 & 64.9 & \underline{64.5} & 60.4 & 64.8 \\
        Eagle2.5-8B & 8B & 86.4 & 27.3 & 10.1 & 50.9 & 56.1 & 40.4 & 51.0 & 46.0 \\
        Penguin-VL-8B & 8B & 88.2 & 74.7 & 41.8 & 58.1 & 58.3 & 54.0 & 43.7 & 59.8 \\
        Molmo2-8B & 8B & 87.6 & 73.4 & 41.6 & 63.2 & 62.7 & 56.2 & 60.4 & 63.6 \\
        Perception-LM-8B & 8B & 84.6 & 69.4 & 37.2 & 42.3 & 45.2 & 42.2 & 23.1 & 49.1 \\
        Ovis2.5-9B & 9B & 88.2 & 77.2 & 44.3 & 62.0 & 59.2 & 43.6 & 48.3 & 60.4 \\
        MiniCPM-V-4.5 & 9B & 87.1 & 73.8 & 43.5 & 57.5 & 60.1 & 56.0 & 52.3 & 61.5 \\
        GLM-4.6V-Flash & 9B & 82.4 & 63.4 & 42.2 & 14.0 & 6.9 & 13.8 & 53.0 & 39.4 \\
        Llama-3.2-11B-Vision & 11B & 83.4 & 60.8 & 30.3 & 31.4 & 26.9 & 20.7 & 30.2 & 40.5 \\
        Long-VITA-14B (128K) & 14B & 83.3 & 57.3 & 35.4 & 67.0 & 59.6 & 51.3 & 55.9 & 58.5 \\
        Kimi-VL-A3B-Instruct & 16B & 87.9 & 62.5 & 38.4 & 60.7 & 61.0 & 54.2 & 55.6 & 60.0 \\
        \midrule
        \rowcolor{TableGroup}
        \multicolumn{10}{l}{\textcolor{TableText}{\textbf{Reasoning-oriented MLLMs}}} \\
        Vision-R1-7B & 7B & 85.6 & 60.1 & 39.2 & 57.8 & 55.7 & 46.2 & 36.8 & 54.5 \\
        PRCO-7B & 7B & 85.0 & 69.9 & 40.3 & 63.8 & 61.8 & 52.5 & 51.9 & 60.7 \\
        Phi-4-reasoning-vision-15B & 15B & 81.1 & 72.5 & 43.1 & 49.1 & 61.8 & 52.1 & 47.0 & 58.1 \\
        \midrule
        \rowcolor{TableGroup}
        \multicolumn{10}{l}{\textcolor{TableText}{\textbf{Document-specialized models}}} \\
        SciMDR-7B & 7B & 86.3 & 75.6 & 37.9 & 68.6 & 58.8 & 47.3 & 49.1 & 60.5 \\
        DocSeeker-7B & 7B & 85.7 & 74.0 & 38.4 & 68.3 & 57.0 & 50.5 & 45.2 & 59.9 \\
        \midrule
        \rowcolor{TableGroup}
        \multicolumn{10}{l}{\textcolor{TableText}{\textbf{Task-adapted backbones}}} \\
        Vision-R1-7B + SFT & 7B & 85.4 & 66.2 & 43.1 & 59.6 & 57.5 & 47.7 & 42.3 & 57.4 \\
        Vision-R1-7B + SFT + GRPO & 7B & 86.9 & 68.5 & 43.7 & 60.8 & 59.2 & 49.7 & 48.3 & 59.6 \\
        Qwen3-VL-8B-Instruct + SFT & 8B & 87.6 & 75.4 & 42.9 & 73.8 & 65.3 & 61.8 & 41.7 & 64.1 \\
        Qwen3-VL-8B-Instruct + SFT + GRPO & 8B & 88.2 & \underline{77.8} & 43.9 & 74.1 & \underline{67.9} & 63.5 & 45.6 & 65.9 \\   
        \midrule
        \rowcolor{TableOurs}
        \textbf{Ours (Vision-R1-7B)} & 7B
        & \underline{89.8} & 74.9 & \underline{45.2}
        & \underline{74.6} & 66.2 & 62.7
        & \underline{75.8} & \underline{69.9} \\
        \rowcolor{TableOurs}
        \textbf{Ours (Qwen3-VL-8B-Instruct)} & 8B
        & \textbf{90.4} & \textbf{79.5} & \textbf{51.5}
        & \textbf{81.5} & \textbf{72.4} & \textbf{70.2}
        & \textbf{79.7} & \textbf{75.0} \\
        \bottomrule
    \end{tabular}
    }
\end{table*}

\textbf{Paired Diagnostic Feedback.}
Although the shared reward drives joint optimization, answer-level performance alone does not reveal whether further improvements should target evidence acquisition or latent guidance. We therefore introduce two complementary diagnostics that isolate these two potential sources of improvement.

\textit{(i) Retrieval Potential Feedback (RPF).}
RPF identifies questions for which improved evidence acquisition could benefit the current perception--reasoning pipeline. With the Perceiver and Reasoner fixed, we replace the retrieved evidence $E$ with annotated evidence $E^*$ and recompute the corresponding latent ontology $Z^*=C_{\psi}(E^*,q)$. Let $S^{\mathrm{gold}}(q)$ and $S^Z(q)$ denote the rubric scores of freshly generated responses using annotated and retrieved evidence, respectively, with their corresponding latent representations. The retrieval potential is defined as
\begingroup
\small
\begin{equation}
    d_{\mathrm{ret}}(q)
    =\bigl[S^{\mathrm{gold}}(q)-S^Z(q)\bigr]_+,
    \label{eq:am_rpf}
\end{equation}
\endgroup
where $[x]_+=\max(0,x)$. A positive gap indicates that more complete or better-aligned evidence could improve answering under the current Perceiver and Reasoner. Questions with positive RPF scores are added to the RPF pool for subsequent Retriever updates.

\textit{(ii) Latent Efficacy Feedback (LEF).}
In contrast, LEF assesses whether the current latent ontology effectively supports reasoning over the retrieved evidence. Keeping the evidence and question fixed, we compare responses generated with and without latent guidance. Let $S^0(q)$ denote the rubric score of the response generated without $Z$, while $S^Z(q)$ denotes the corresponding score with $Z$. We define the latent efficacy gap as
\begingroup
\small
\begin{equation}
    d_{\mathrm{use}}(q)
    =\bigl[S^0(q)-S^Z(q)\bigr]_+.
    \label{eq:am_lef}
\end{equation}
\endgroup
A positive gap identifies questions where latent guidance reduces answering performance despite unchanged evidence access. These questions enter the LEF pool for subsequent Perceiver and Reasoner updates. The stability of both diagnostic pools is analyzed in Appendix~\ref{apx:supplementary_analysis_experiment}.

\textbf{Diagnostic-Guided Resampling.}
After each training cycle, we apply RPF and LEF to the questions encountered in that cycle to construct feedback pools for the next. Subsequent batches combine regular and diagnostic-selected questions, using RPF for Retriever updates and LEF for Perceiver and Reasoner updates. All diagnostics use training data, and selected questions receive fresh trajectories from the current model. This strategy prioritizes questions that require further improvement in retrieval or latent guidance.

\begin{table*}[t]
    \centering
    \caption{Ablations of latent perception, co-evolution, and diagnostic feedback on Vision-R1-7B. Removing the Perceiver also omits its supervised initialization. Freezing applies only during co-evolution.
    }
    \label{tab:am_ablation}
    \small
    \setlength{\tabcolsep}{4pt}
    \renewcommand{\arraystretch}{1.12}
    \scalebox{0.85}{%
    \begin{tabular}{@{}l *{8}{c}@{}}
        \toprule
        \textbf{Variant}
        & \textbf{ChartQA}
        & \textbf{CharXiv-D}
        & \textbf{CharXiv-R}
        & \textbf{SPIQA-A}
        & \textbf{SPIQA-B}
        & \textbf{SPIQA-C}
        & \textbf{SciMDR-Eval}
        & \textbf{Avg.} \\
        \midrule
        \rowcolor{TableOurs}
        \textbf{Ours (on Vision-R1-7B)}
        & \textbf{89.8} & \textbf{74.9} & \textbf{45.2}
        & \textbf{74.6} & \textbf{66.2} & \textbf{62.7}
        & \textbf{75.8} & \textbf{69.9} \\
        \midrule

        \multicolumn{9}{l}{
            \textit{(a) Latent perception and initialization}} \\
        w/o Perceiver
        & 85.0 & 65.2 & 36.0 & 68.6 & 60.1 & 52.1 & 58.7 & 60.8 \\
        w/o Retriever initialization
        & 84.2 & 62.0 & 35.8 & 61.1 & 57.5 & 49.1 & 48.1 & 56.8 \\
        w/ Reranker-only Top-$K$
        & 85.7 & 60.3 & 37.2 & 62.6 & 59.2 & 47.9 & 46.5 & 57.1 \\
        \midrule

        \multicolumn{9}{l}{
            \textit{(b) Co-evolution: which components adapt and how}} \\
        w/o co-evolution
        & 85.4 & 64.6 & 37.2 & 71.9 & 59.6 & 51.1 & 63.5 & 61.9 \\
        Freeze Perceiver and Retriever
        & 87.2 & 63.2 & 40.7 & 66.2 & 59.6 & 50.7 & 67.7 & 62.2  \\
        Freeze Perceiver
        & 86.4 & 67.3 & 38.8 & 68.2 & 63.6 & 55.2 & 68.5 & 64.0 \\
        Freeze Retriever
        & 88.5 & 66.8 & 39.6 & 67.6 & 61.8 & 54.6 & 71.2 & 64.3 \\
        Joint updates (no alternation)
        & 87.5 & 65.4 & 40.1 & 67.1 & 62.3 & 52.9 & 71.5 & 63.8 \\
        \midrule
        \multicolumn{9}{l}{
            \textit{(c) Diagnostic feedback for targeted learning}} \\
        w/o RPF
        & 89.2 & 69.1 & 42.9 & 70.6 & 64.5 & 61.4 & 73.0 & 67.2  \\
        w/o LEF
        & 88.7 & 68.2 & 43.3 & 71.1 & \textbf{66.2} & 61.1 & 74.1 & 67.5  \\
        w/o RPF and LEF
        & 87.8 & 66.4 & 43.8 & 69.5 & 64.0 & 60.4 & 73.6 & 66.5 \\
        w/ Random Resampling
        & 86.4 & 62.1 & 44.3 & 68.6 & 64.9 & 60.4 & 
        71.6 & 65.5 \\
        \midrule
        \multicolumn{9}{l}{
        \textit{(d) Perceiver architecture and query conditioning}} \\
        w/o QueryBuilder
        & 85.7 & 64.3 & 39.8 & 66.5 & 58.3 & 55.7 & 66.9 & 62.5 \\
        Question-only QueryBuilder
        & 86.4 & 67.8 & 41.7 & 67.5 & 59.7 & 59.5 & 70.4 & 64.7 \\
        Evidence-only QueryBuilder
        & 87.3 & 68.5 & 42.3 & 68.6 & 59.3 & 56.3 & 68.7 & 64.4 \\
        \bottomrule
        
    \end{tabular}%
    }
\end{table*}

\subsection{Inference}
\label{sec:am_implementation}

Given a scientific document $D$ and a question $q$, our framework constructs a source-aligned candidate pool $\mathcal C(D)$, from which the Retriever sequentially selects and packs evidence $E$. The Perceiver then constructs the question-conditioned latent ontology $Z=C_{\psi}(E,q)$. Finally, the Reasoner jointly processes $E$, $Z$, and $q$ to autoregressively generate the reasoning trace and final answer, retaining direct access to original evidence alongside latent guidance.

\section{Experiment}

\subsection{Experimental Setup}

\paragraph{Datasets and Metrics.} Following SciMDR~\citep{chen2026scimdr}, we evaluate our method on four benchmarks: ChartQA~\citep{masry2022chartqa}, CharXiv-D/R~\citep{wang2024charxiv}, SPIQA-A/B/C~\citep{pramanick2024spiqa}, and SciMDR-Eval~\citep{chen2026scimdr}. Following~\citet{chen2026scimdr}, we adopt standard evaluation protocols and metrics provided by each benchmark.

\textbf{Baselines.}
We compare our method with baselines organized into four complementary groups according to their primary evaluation role.
(1) General-purpose MLLMs cover diverse multimodal architectures, including Qwen2.5-VL-7B~\citep{Qwen2.5-VL}, Qwen3-VL-8B-Instruct~\citep{bai2025qwen3}, Keye-VL-1.5-8B~\citep{yang2025kwai}, LLaVA-OneVision-1.5-8B~\citep{an2025llava}, InternVL3-8B~\citep{zhu2025internvl3}, InternVL3.5-8B~\citep{wang2025internvl3}, Eagle2.5-8B~\citep{chen2026eagle}, Penguin-VL-8B~\citep{zhang2026penguin}, Molmo2-8B~\citep{clark2026molmo2}, Perception-LM-8B~\citep{cho2026perceptionlm}, Ovis2.5-9B~\citep{lu2025ovis2}, MiniCPM-V-4.5~\citep{yao2024minicpm}, GLM-4.6V-Flash~\citep{hong2025glm}, Llama-3.2-11B-Vision, and Kimi-VL-A3B-Instruct~\citep{team2025kimi}, Long-VITA-14B~\citep{shen2025long}.
\textbf{(2) Reasoning-oriented MLLMs}, namely Vision-R1-7B~\citep{huang2026vision}, PRCO-7B~\citep{miao2026seeing}, and Phi-4-reasoning-vision-15B~\citep{aneja2026phi}. (3) Document-specialized models include the 7B model fine-tuned on SciMDR~\citep{chen2026scimdr} and DocSeeker-7B~\citep{yan2026docseeker}. 
\textbf{(4) Task-adapted backbones} include SFT and SFT + GRPO, trained on the same 7,000 SciMDR examples used by our method.
More implementation details are in Appendix~\ref{sec:am_appendix_baselines}.

\textbf{Training Details.}
We use Vision-R1-7B~\citep{huang2026vision} and Qwen3-VL-8B-Instruct~\citep{bai2025qwen3} as the multimodal backbone and train on 7,000 examples from the SciMDR training set. The Retriever uses Qwen3-VL-Reranker-2B~\citep{li2026qwen3} to encode question--candidate pairs, with its parameters frozen throughout training. Following supervised initialization, we perform three co-evolution cycles with 20 optimization steps per component per cycle. Each update uses a batch of 32. The first cycle uses regular sampling only; subsequent batches combine 24 regularly sampled questions with 8 questions from the relevant feedback pool, with any shortfall filled by regular sampling. The maximum input length is 128K tokens, and the reward weights are $\lambda_{\mathrm{ret}}$=$\lambda_{\mathrm{gain}}$=0.5. More training details are in Appendix~\ref{sec:am_appendix_training}.

\begin{table}[t]
    \centering
    \caption{Retriever performance on Vision-R1-7B. Visual evidence includes figures and tables. SPIQA visual metrics are averaged over A/B/C, while text F1 is averaged over B/C.}
    \label{tab:am_retriever_analysis}
    \small
    \setlength{\tabcolsep}{3pt}
    \renewcommand{\arraystretch}{1.16}

    \scalebox{0.76}{%
    \begin{tabular}{@{}l cc cc c c@{}}
        \toprule
        \textbf{Training Status}
        & \multicolumn{2}{c}{
            \shortstack{\textbf{SPIQA}\\ \textbf{Visual}}}
        & \multicolumn{2}{c}{
            \shortstack{\textbf{SciMDR-Eval}\\ \textbf{Visual}}}
        & \shortstack{\textbf{SPIQA}\\ \textbf{Text}}
        & \shortstack{\textbf{SciMDR-Eval}\\ \textbf{Text}} \\
        \cmidrule(lr){2-3}
        \cmidrule(lr){4-5}
        \cmidrule(lr){6-6}
        \cmidrule(lr){7-7}
        & Hit Rate & F1 & Hit Rate & F1 & F1 & F1 \\
        \midrule

        Before Training
        & 48.94 & 43.51 & 56.74 & 48.30 & 22.46 & 23.62 \\

        After initialization
        & 56.59 & 54.41 & 78.37 & 72.02 & 33.10 & 34.57 \\

        \rowcolor{TableOurs}
        \textbf{After co-evolution}
        & \textbf{61.71} & \textbf{60.89}
        & \textbf{80.85} & \textbf{79.74}
        & \textbf{39.01} & \textbf{40.14} \\

        \bottomrule
    \end{tabular}%
    }
\end{table}

\subsection{Main Results}
\label{sec:am_main_results}

\paragraph{Consistent Gains Across Scientific Reasoning Benchmarks.}
Our framework consistently outperforms strong baselines across both backbones, achieving average gains of 9.1 and 4.0 points with Qwen3-VL-8B-Instruct and Vision-R1-7B, respectively. The consistent gains across both backbones demonstrate the effectiveness of jointly learning evidence acquisition, cross-modal perception, and grounded reasoning. Notably, even with the 7B Vision-R1 backbone, our method surpasses all compared models with 11B--16B backbones. We further analyze inference cost and hyperparameter sensitivity in Appendix~\ref{sec:am_appendix_efficiency}.

\begin{table}[t]
    \centering
    \caption{Answering performance under latent-input interventions on Vision-R1-7B. }
    \label{tab:am_z_intervention}
    \small
    \setlength{\tabcolsep}{3pt}
    \renewcommand{\arraystretch}{1.16}
    \scalebox{0.80}{%
    \begin{tabular}{l *{5}{c}}
        \toprule
        \textbf{Condition} & \textbf{SPIQA-A} & \textbf{SPIQA-B}
        & \textbf{SPIQA-C} & \textbf{SciMDR-Eval} & \textbf{Avg.} \\
        \midrule
        \rowcolor{TableOurs}
        Full $Z$ & \textbf{74.6} & \textbf{66.2} & \textbf{62.7}
        & \textbf{75.8} & \textbf{69.8} \\
        No $Z$ & 68.5 & 59.2 & 54.6 & 62.1 & 61.1 \\
        Random $Z$ & 50.2 & 41.7 & 36.5 & 45.3 & 43.4 \\
        Mismatched $Z$ & 52.6 & 43.9 & 40.3 & 47.6 & 46.1 \\
        \bottomrule
    \end{tabular}
    }
\end{table}

\begin{table}[t]
\centering
\caption{Perceiver input-modality interventions on SciMDR-Eval.}
\label{tab:am_cross_modal_analysis}
\small
\setlength{\tabcolsep}{4pt}
\renewcommand{\arraystretch}{1.15}
\scalebox{0.85}{%
\begin{tabular}{cccc}
\toprule
\textbf{Full $Z$} & \textbf{Text-only $Z$} & \textbf{Visual-only $Z$} & \textbf{No $Z$} \\
\midrule
\textbf{75.8} & 65.8 & 67.3 & 62.1 \\
\bottomrule
\end{tabular}
}
\end{table}

\subsection{Ablation Studies}
\label{sec:am_ablation}

\textbf{Component Contributions and Diagnostic Co-Evolution.}
Table~\ref{tab:am_ablation} validates the contributions of individual components and their coordinated optimization. Removing the Perceiver substantially degrades performance despite the Reasoner's direct access to original evidence, highlighting the value of complementary latent guidance. Retriever initialization and learned source selection also contribute substantially, with the latter outperforming budget-matched reranker-only Top-$K$ retrieval. QueryBuilder ablations further confirm the importance of jointly conditioning latent ontology construction on the question and retrieved evidence. Beyond individual components, alternating co-evolution outperforms simultaneous joint updates and variants with frozen components, demonstrating the benefit of coordinated adaptation. Moreover, RPF and LEF provide complementary benefits: removing either degrades performance, while disabling both or replacing diagnostic-guided resampling with uniform resampling leads to further performance drops. These results highlight the importance of paired diagnostics in directing component-specific learning throughout co-evolution.

\textbf{Co-Evolving Evidence Acquisition.}
Table~\ref{tab:am_retriever_analysis} shows consistent improvements across all six retrieval metrics from supervised initialization to the final co-evolved model. This indicates that the Retriever progressively improves source selection as the Perceiver and Reasoner evolve. Moreover, the evidence-usage analysis in Appendix~\ref{sec:am_appendix_evidence_usage} shows that these gains do not require exhausting the retrieval budget: the trained Retriever maintains average source counts and token usage below the prescribed limits while improving localization of textual and visual evidence.

\subsection{Analysis Studies}

\paragraph{Latent Ontology as Complementary Reasoning Guidance.}
We investigate whether the latent ontology provides useful guidance beyond the Reasoner's direct access to original evidence. All model parameters and the Reasoner's textual and visual inputs are held fixed across interventions. As shown in Table~\ref{tab:am_z_intervention}, removing $Z$ reduces performance, while replacing it with random or mismatched representations causes substantially larger degradation. This indicates that effective latent guidance depends on its alignment with the current question and retrieved evidence. We further examine the role of cross-modal information in constructing $Z$ on SciMDR-Eval, where questions require both textual and visual evidence. As shown in Table~\ref{tab:am_cross_modal_analysis}, restricting the Perceiver to either textual or visual inputs degrades performance, even though the Reasoner retains access to both modalities. These findings support the role of the Perceiver in organizing complementary multimodal evidence beyond direct source access; intervention details in Appendix~\ref{sec:am_latent_interventions}.

\begin{table}[t]
    \centering
\caption{Co-evolution dynamics on SciMDR-Eval with Vision-R1-7B. Cycle 0 denotes supervised initialization; Cycle 4 is an additional training checkpoint.}
    \label{tab:am_coevolution_cycles}
    \small
    \setlength{\tabcolsep}{3pt}
    \renewcommand{\arraystretch}{1.16}
    \scalebox{1}{%
    \begin{tabular}{l *{5}{c}}
        \toprule
        \textbf{Cycle} & w/o $Z$ & w/ $Z$ & $\Delta$Score & Text F1 & Visual F1 \\
        \midrule
        0 & 57.3 & 63.5 & 6.2 & 34.57 & 72.02 \\
        1 & 58.8 & 70.3 & 11.5 & 36.14 & 74.39 \\
        2 & 62.7 & 72.5 & 9.8 & 38.27 & 77.06 \\
        \rowcolor{TableOurs}
        \textbf{3} & 62.1 & \textbf{75.8} & \textbf{13.7} & 40.14 & \textbf{79.74} \\
        4 & \textbf{64.9} & 74.2 & 9.3 & \textbf{41.55} & 78.33 \\
        \bottomrule
    \end{tabular}}
\end{table}

\textbf{Co-Evolution Dynamics and Self-Improvement.}
Table~\ref{tab:am_coevolution_cycles} tracks answering performance, retrieval quality, and the contribution of latent guidance across training cycles. From initialization to cycle 3, the with-$Z$ score increases from 63.5 to 75.8, accompanied by improvements in both text and visual retrieval F1. These gains indicate that co-evolution progressively improves evidence acquisition and its utilization for reasoning. However, the improvements are not monotonic: at cycle 4, text F1 and no-$Z$ performance continue to increase, while visual F1 declines and the with-$Z$ score drops to 74.2. This divergence highlights the interdependence of retrieval and latent perception: improvements in individual components do not necessarily translate into better end-to-end reasoning.

\section{Conclusion}
\label{sec:am_conclusion}

We present an agentic framework that synergizes evidence acquisition and utilization for multimodal scientific document reasoning. By combining relation-aware retrieval with question-conditioned latent ontology construction, our framework enables the Reasoner to integrate complementary multimodal information while retaining direct access to original sources. To further improve these interdependent capabilities, we introduce diagnostic-guided co-evolution, which coordinates component updates through a shared answering reward and leverages paired diagnostics to guide targeted learning. Experiments across four benchmarks and seven evaluation settings demonstrate consistent improvements over strong baselines, with gains of 4.0 and 9.1 points using Vision-R1-7B and Qwen3-VL-8B-Instruct.


{
    \small
    \bibliographystyle{ieeenat_fullname}
    \bibliography{main}
}
\clearpage
\appendix
\section{Appendix}
\subsection{Additional Implementation Details}
\label{sec:am_appendix_training}

\paragraph{Training details.}
We train two backbone variants, Vision-R1-7B and Qwen3-VL-8B-Instruct, on NVIDIA A800 (80\,GB) GPUs using 7,000 examples from the SciMDR training set~\citep{chen2026scimdr}. We compare the figures and tables in the training documents against those in the evaluation sets and find no duplicates. Within each variant, the Perceiver and Reasoner share a frozen backbone with separate low-rank adaptation (LoRA) adapters of rank $r=16$ and scaling parameter $\alpha=32$. The Perceiver uses $N_q=32$ query slots and $M=32$ output slots, projecting its outputs into the Reasoner's embedding space. The latent sequence $Z$ is appended after the selected evidence, which remains directly accessible to the Reasoner. Retrieval is limited to $K_{\max}=4$ source units and $B_E=6{,}144$ evidence tokens, including native visual tokens. A frozen Qwen3-VL-Reranker-2B~\citep{li2026qwen3} supplies question--candidate features. Perceiver initialization runs for two epochs with a global batch size of 64, using annotated evidence in place of retrieved evidence. Reference-response loss gradients propagate through the frozen Reasoner to update only the Perceiver. The Retriever policy is initialized separately for two epochs with batches of 32 questions. We then run three co-evolution cycles, selected by the highest mean answer score on a held-out 500-example validation set. Each cycle updates the Perceiver, Retriever, and Reasoner in that order for 20 optimization steps each, keeping the other components fixed during each phase. Each step uses 32 questions and $G=8$ fresh retrieval trajectories per question. For each evidence set, the Reasoner generates paired with-$Z$ and no-$Z$ responses using temperature 0.6, top-$p=0.95$, and a maximum of 4,096 generated tokens per response. The shared reward assigns unit weight to the with-$Z$ answer score and uses $\lambda_{\mathrm{ret}}=\lambda_{\mathrm{gain}}=0.5$. 

After each cycle, we apply the paired diagnostics to the training questions encountered in that cycle using the updated model. For each question, we use one retrieved evidence set and generate one fresh response per condition: retrieved evidence with $Z$, the same evidence without $Z$, and annotated evidence with recomputed $Z^{*}$. The annotated-evidence condition follows the same packing rules and source-count and token budgets as the retrieved-evidence condition. RPF and LEF share the retrieved with-$Z$ response and its rubric score. These single-response diagnostics are separate from the $G=8$ training rollouts. Questions with positive diagnostic scores enter the corresponding RPF or LEF pool for the next cycle. The first cycle uses regular sampling; subsequent batches combine 24 regular questions with up to 8 feedback questions, drawn from RPF for the Retriever and LEF for the Perceiver and Reasoner. Regular sampling fills any shortfall. Each question can be drawn from a feedback pool at most once per component per cycle, so a LEF question remains available to the Reasoner after being used by the Perceiver.

For the task-adapted backbone baselines, we fine-tune each released backbone on the same 7,000 training examples for two epochs with a global batch size of 64 to obtain the SFT baseline. The SFT + GRPO variant starts from the corresponding SFT checkpoint and undergoes 180 optimization steps using rubric-based answer rewards, with 32 questions per batch and $G=8$ rollouts per question. This matches the total update count of our co-evolution training, with the same decoding settings. To ensure comparable evidence access, both baselines receive the question and evidence selected by our Retriever after supervised initialization. This Retriever remains frozen throughout baseline training and evaluation, with the same evidence packing, and retrieval budgets as our full model.

\begin{table*}[t]
    \centering
    \caption{Inference quality and cost on SciMDR-Eval. All variants of our framework use Vision-R1-7B. Values are per-question averages.}
    \label{tab:am_inference_efficiency}
    \small
    \setlength{\tabcolsep}{4pt}
    \renewcommand{\arraystretch}{1.15}
    \scalebox{1}{%
    \begin{tabular}{l *{7}{c}}
        \toprule
        \multirow{2}{*}{\textbf{Method}}
        & \multirow{2}{*}{\textbf{Score}$\uparrow$}
        & \multirow{2}{*}{\shortstack{\textbf{Output}\\\textbf{tokens}}}
        & \multicolumn{5}{c}{\textbf{Latency (s / question)}$\downarrow$} \\
        \cmidrule(lr){4-8}
        & & & Prep.
        & Retrieval
        & Perceiver
        & Reasoner
        & Total \\
        \midrule
        Vision-R1-7B
        & 36.8 & 553 & -- & -- & -- & 35.58 & 35.58 \\
        InternVL3.5-8B
        & 60.4 & 470 & -- & -- & -- & 15.92 & 15.92 \\
        \midrule
        Ours w/ Reranker-only Top-$K$
        & 46.5 & 522 & 0.12 & 2.68 & 0.47 & 18.40 & 21.67 \\
        Ours w/o Perceiver
        & 58.7 & 547 & 0.12 & 1.14 & -- & 17.28 & 18.54 \\
        \rowcolor{TableOurs}
        \textbf{Ours (full)}
        & 75.8 & 461 & 0.12 & 2.37 & 0.61 & 17.66 & 20.76 \\
        \bottomrule
    \end{tabular}}
\end{table*}

\textbf{Ablation settings.}
The joint-training and frozen-component variants in Table~\ref{tab:am_ablation} share the full model's supervised initialization, rollout configuration, and optimization settings. Cyclic and joint training each run for 180 optimization steps, but allocate updates differently: cyclic training updates each component 60 times, whereas joint training updates all three components at every step. The joint-training variant uses neither RPF nor LEF. Frozen components retain their post-initialization parameters throughout co-evolution, while the remaining components, including the Reasoner, continue to train. The random-resampling control replaces feedback-selected questions with an equal number sampled uniformly from the preceding cycle's training questions. All co-evolution variants generate fresh rollouts for each update. The three QueryBuilder ablations retain the Latent Evidence Consolidator and the Reasoner's access to the question and original evidence. Removing the QueryBuilder replaces its generated queries with $N_q=32$ learnable query vectors shared across examples. The question-only and evidence-only variants restrict its input to $H_q$ and $H_E$, respectively. In each case, the Consolidator still processes the full retrieved evidence and produces $M=32$ latent vectors, following the same supervised initialization and co-evolution schedules as the full model. The reranker-only Top-$K$ variant replaces the learned retrieval policy with selection by frozen reranker scores throughout training and inference. Retriever initialization and policy updates are omitted; Perceiver initialization remains unchanged, and subsequent training updates only the Perceiver and Reasoner using reranker-selected evidence. In Table~\ref{tab:am_retriever_analysis}, ``Before Training'' denotes budget-constrained Top-$K$ selection by the frozen reranker without a learned retrieval policy. Visual Hit Rate measures the percentage of evaluated questions for which the selected evidence contains at least one annotated figure or table. Text and visual F1 follow the source-aligned definitions in Appendix~\ref{sec:am_appendix_evidence}, and SPIQA metrics are averaged over A/B/C.

\paragraph{Latent-input and modality interventions.}
\label{sec:am_latent_interventions} 
All interventions use the trained Vision-R1-7B model with parameters fixed. For Table~\ref{tab:am_z_intervention}, Random $Z$ replaces the learned latent sequence with randomly initialized vectors of the same number and dimension. Mismatched $Z$ uses the representation generated by the same Perceiver from another question and its annotated evidence. For Table~\ref{tab:am_cross_modal_analysis}, Text-only $Z$ retains textual evidence, including figure and table captions, but removes image inputs. Visual-only $Z$ retains figure and table images but excludes captions and other evidence text. These changes are applied to the Perceiver's evidence inputs before feature extraction, while its question remains unchanged. Across all conditions, the Reasoner receives the same original multimodal evidence and question. For SciMDR-Eval, we follow the official evaluation protocol~\citep{chen2026scimdr} using GPT-5-mini, which is distinct from the judge used to compute training-time rewards.

\paragraph{Baseline Implementation Details}
\label{sec:am_appendix_baselines}

We use publicly released weights for all general-purpose and reasoning-oriented MLLM baselines. For SciMDR-7B, we follow the evaluation setup described in the original paper~\citep{chen2026scimdr}. DocSeeker-7B is the baseline that we reproduce and train on our training corpus: we implement its published method~\citep{yan2026docseeker} and train it on the same 7,000 examples used for our framework.

\begin{table}[t]
    \centering
    \caption{Evidence usage for questions with complete source content ($K_{\max}=4$, $B_E=6{,}144$) on Vision-R1-7B. Values are per-question means.}
    \label{tab:am_inference_evidence_usage}
    \small
    \setlength{\tabcolsep}{3pt}
    \renewcommand{\arraystretch}{1.16}
    \scalebox{0.95}{%
    \begin{tabular}{l *{5}{c}}
        \toprule
        \textbf{Metric} & \textbf{SPIQA-A} & \textbf{SPIQA-B} & \textbf{SPIQA-C} & \textbf{SciMDR-Eval} & \textbf{Avg.} \\
        \midrule
        Sources & 3.09 & 2.68 & 2.26 & 2.56 & 2.65 \\
        Tokens & 1461 & 2230 & 1020 & 2625 & 1834 \\
        \bottomrule
    \end{tabular}
    }
\end{table}

\subsection{Policy Optimization for Cyclic Co-Evolution}
\label{sec:am_appendix_optimization}

We use GRPO~\citep{shao2024deepseekmath} to optimize the active component under the shared reward in Section~\ref{sec:am_stage3}. Within each valid group of rollouts for the same question, rewards are normalized into detached advantages $A_i=(R_i-\mu_R)/(\sigma_R+\eta)$, where $\mu_R$ and $\sigma_R$ are the group reward mean and standard deviation, and $\eta>0$ ensures numerical stability. For active block $b\in\{\phi,\psi,\theta\}$, we define the probability ratio $\rho_{i,t}^{b}=\pi_b(a_{i,t}\mid s_{i,t})/\pi_{b,\mathrm{old}}(a_{i,t}\mid s_{i,t})$ using the recorded rollout-policy probabilities, and its clipped counterpart $\bar\rho_{i,t}^{b}=\operatorname{clip}(\rho_{i,t}^{b},1-\epsilon,1+\epsilon)$. The objective is
\begingroup
\small
\begin{equation}
    \begin{aligned}
    \ell_{i,t}^{b}
      &=-\min\{\rho_{i,t}^{b}A_i,\bar\rho_{i,t}^{b}A_i\}
        +\beta k_{i,t}^{b},\\
    \mathcal{L}_{\mathrm{coev}}^{b}
      &=\mathbb{E}_{i}\left[\frac{1}{T_i^b}
        \sum_{t=1}^{T_i^b}\ell_{i,t}^{b}\right].
    \end{aligned}
    \label{eq:am_grpo}
\end{equation}
\endgroup
Here, $T_i^b$ counts valid retrieval actions or generated response tokens, $\epsilon$ is the clipping radius, and $\beta$ weights KL regularization. At each scored position, we use $k^b=r^{\mathrm{ref}}-\log r^{\mathrm{ref}}-1$, where $r^{\mathrm{ref}}=\pi_{b,\mathrm{ref}}/\pi_b$. The reference policy is frozen at the start of co-evolution. For response probabilities, it uses the initial Perceiver and Reasoner, with the reference latent representation computed from the same retrieved evidence. The rollout policy is refreshed for each batch, while its recorded probabilities remain fixed during optimization. Losses are averaged over valid positions within each trajectory and then over trajectories.

For Retriever updates, $\pi_b$ denotes the retrieval-action distribution. For Perceiver and Reasoner updates, it denotes the autoregressive response distribution conditioned on latent guidance:
$p_{\theta}(y\mid E,q,C_{\psi}(E,q)).$
During Perceiver updates, the Reasoner's parameters $\theta$ remain fixed, while gradients propagate through the Reasoner to the latent representation $Z$ and Perceiver parameters $\psi$. Thus, GRPO optimizes the Perceiver through its effect on the induced response distribution. During Reasoner updates, the Perceiver remains fixed and only $\theta$ is optimized. The shared backbone and reranker remain frozen throughout co-evolution.

For each batch, the old policy probabilities are recorded from the policy that generates the rollouts and remain fixed during optimization. The current policy probabilities are recomputed using the updated active component. The reference policy is frozen at the start of co-evolution. For retrieval actions, it uses the initial Retriever; for response distributions, it uses the initial Perceiver and Reasoner, with the reference latent representation computed from the same retrieved evidence. RPF and LEF determine which questions are revisited; they do not modify the GRPO objective or replace the shared reward.

\subsection{Supplementary Analysis Experiments}
\label{apx:supplementary_analysis_experiment}

\paragraph{Evidence Usage at Inference.}
\label{sec:am_appendix_evidence_usage}
Table~\ref{tab:am_inference_evidence_usage} summarizes evidence usage for questions with complete source content. With $K_{\max}=4$ and $B_E=6{,}144$, the mean number of selected sources is below four on every dataset, and mean evidence-token counts range from 1,020 to 2,625. Thus, retrieval uses less than the available capacity on average. Together with the text and visual F1 improvements in Table~\ref{tab:am_retriever_analysis}, these observations support selective evidence acquisition that combines limited evidence usage with improved source grounding for subsequent perception and reasoning.

\paragraph{Inference Cost Analysis.}
\label{sec:am_appendix_efficiency}
We evaluate the quality--cost trade-off of our framework with Vision-R1-7B using complete documents as input. All methods run under the same NVIDIA A800 hardware configuration, batch size of one. Output tokens include reasoning and the final answer. Prep.\ includes document loading, parsing, and structured evidence-candidate construction; retrieval includes reranker scoring and policy selection; Reasoner time includes prefill and decoding. Total latency is measured from document input to answer completion, with models preloaded and without document caches. Single-pass baselines process the full document directly, without candidate construction or retrieval.

Table~\ref{tab:am_inference_efficiency} shows that the full model improves answer quality over single-pass Vision-R1-7B with 41.7\% lower total latency. It also outperforms reranker-only Top-$K$ with slightly lower latency. The Perceiver takes 0.61\,s per question, accounting for 2.9\% of total latency, while the full model outperforms the no-Perceiver variant with fewer output tokens. Compared with InternVL3.5-8B, our method gains 15.4 score points at an additional 4.84\,s per question. These results support allocating computation to adaptive evidence acquisition and latent perception without requiring longer responses.

\paragraph{Sensitivity to Reward Weights.}
\label{sec:am_appendix_reward_sensitivity}
To assess how reward weighting affects the co-evolution of evidence acquisition and use, we vary $\lambda_{\mathrm{ret}}$ and $\lambda_{\mathrm{gain}}$ individually over $\{0, 0.25,0.5,0.75\}$, keeping the other weight fixed at 0.5. All experiments use Vision-R1-7B and start from the same supervised initialization, with three co-evolution cycles, $G=8$ rollouts per question, and all other settings unchanged. We reserve an additional 500 examples from the SciMDR training split, disjoint from the 7,000 training examples, as a validation set for comparing reward configurations. These validation examples are excluded from parameter updates and RPF/LEF pool construction. Table~\ref{tab:am_reward_sensitivity} reports performance on SPIQA-A/B/C and SciMDR-Eval to examine how the reward choices affect benchmark performance; these evaluation results are not used for hyperparameter selection. We select $\lambda_{\mathrm{ret}}=\lambda_{\mathrm{gain}}=0.5$ for experiments, as this configuration achieves the best performance on the held-out validation set among the tested settings.

\begin{table}[t]
\centering
\caption{Reward-weight sensitivity on Vision-R1-7B. }
\label{tab:am_reward_sensitivity}
\small
\setlength{\tabcolsep}{3pt}
\renewcommand{\arraystretch}{1.15}
\scalebox{0.82}{%
\begin{tabular}{cc cccc}
\toprule
$\lambda_{\mathrm{ret}}$
& $\lambda_{\mathrm{gain}}$
& \textbf{SPIQA-A}
& \textbf{SPIQA-B}
& \textbf{SPIQA-C}
& \textbf{SciMDR-Eval} \\
\midrule
\rowcolor{TableOurs}
0.50 & 0.50 & 74.6 & \textbf{66.2} & \textbf{62.7} & 75.8 \\
\midrule
0.00 & 0.50 & 70.7 & 61.4 & 60.4 & 71.2 \\
0.25 & 0.50 & 73.4 & 64.5 & 61.9 & 73.8 \\
0.75 & 0.50 & 74.3 & 63.6 & 61.1 & \textbf{76.9} \\
\midrule
0.50 & 0.00 & 71.2 & 62.3 & 59.2 & 70.4 \\
0.50 & 0.25 & \textbf{74.9} & 65.4 & 62.3 & 74.2 \\
0.50 & 0.75 & 74.2 & 64.9 & 60.9 & 75.5 \\
\bottomrule
\end{tabular}%
}
\end{table}

\paragraph{Sensitivity to Latent and Query Slot Counts.}
\label{sec:am_appendix_slot_sensitivity}
To examine how the Perceiver's capacity affects evidence use, we vary the output slot count $M$ and QueryBuilder slot count $N_q$ individually over $\{16,32,64\}$, fixing the other at 32. The former controls the number of latent vectors supplied to the Reasoner, while the latter controls the number of queries used to aggregate evidence. Each configuration uses Vision-R1-7B, with the Perceiver initialized at its corresponding slot counts and then trained through the same three-cycle co-evolution schedule. All remaining settings are held fixed. Table~\ref{tab:am_slot_sensitivity} compares the resulting answering performance.

\begin{table}[t]
\centering
\caption{Slot-count sensitivity on Vision-R1-7B. }
\label{tab:am_slot_sensitivity}
\small
\setlength{\tabcolsep}{3pt}
\renewcommand{\arraystretch}{1.15}
\scalebox{0.86}{%
\begin{tabular}{cc cccc}
\toprule
$M$ & $N_q$
& \textbf{SPIQA-A}
& \textbf{SPIQA-B}
& \textbf{SPIQA-C}
& \textbf{SciMDR-Eval} \\
\midrule
\rowcolor{TableOurs}
32 & 32 & 74.6 & \textbf{66.2} & 62.7 & 75.8 \\
\midrule
16 & 32 & 72.7 & 64.5 & \textbf{63.1} & 69.4 \\
64 & 32 & \textbf{75.1} & 61.4 & 62.1 & 73.2 \\
\midrule
32 & 16 & 73.6 & 63.2 & 60.8 & 71.3 \\
32 & 64 & 74.2 & 65.4 & 61.7 & \textbf{76.2} \\
\bottomrule
\end{tabular}%
}
\end{table}

\paragraph{Generation stability of diagnostic feedback.} To assess whether stochastic response generation changes feedback-pool decisions, we randomly sample 150 questions from each of the RPF and LEF pools constructed during training. For each question, we fix the corresponding model checkpoint and original evidence, and independently regenerate three response pairs using the original diagnostic decoding settings. RPF pairs use retrieved and budget-matched annotated evidence with their respective latent representations, while LEF pairs use the same retrieved evidence with and without $Z$. We score the responses with the training-time judge and recompute the pool decision for each pair. The regenerated decisions agree with the original pool-inclusion decisions in 436/450 cases (96.9\%) for RPF and 431/450 cases (95.8\%) for LEF. At the question level, all three decisions agree with the original decision for 139/150 questions (92.7\%) in the RPF pool and 137/150 questions (91.3\%) in the LEF pool. These results indicate that pool inclusion for the selected questions is largely stable under repeated response generation.

\paragraph{Judge Sensitivity of Diagnostic Feedback.}
To assess the sensitivity of diagnostic-guided question selection to the scoring model, we rescore the same training diagnostics with GLM-5.3-Flash~\citep{zeng2026glm}, keeping the generated responses, reference-grounded rubrics, and scoring prompt unchanged. We measure agreement as the proportion of evaluated questions for which both judges make the same pool-inclusion decision. Relative to the original judge, GLM-5.3-Flash achieves agreement rates of 92.4\% for RPF and 93.7\% for LEF, indicating that diagnostic-guided question selection is relatively consistent across the two judges.

\subsection{Evidence Candidate Construction Details}
\label{sec:am_appendix_evidence}

Given an input document $D$, we construct its candidate pool $\mathcal C(D)$ by applying a fixed set of rules to its text, figures, tables, and captions before retrieval. Candidate construction depends only on the document, independently of the question and reference annotations. The same procedure applies to previously unseen documents at inference, without requiring a pre-existing candidate pool. Serialization markers are removed and layout whitespace is normalized, while paragraph boundaries are retained for segmentation. Each text candidate stores its paper and section identifiers and its character interval in the normalized source text. Section headings are retained as context for interpreting selected content.

Text blocks target 384 tokens, with a normal upper limit of 512 tokens including heading context. Within each section, the splitter first prefers paragraph boundaries near the target length and otherwise uses sentence boundaries. It searches for a sentence boundary within the final 64 tokens of a block to provide overlap with the next block; if no such boundary exists, the next block starts at the current endpoint. Exceptionally long ordinary sentences can be split at tokenizer boundaries, with these splits recorded. Display equations and recognized table-like spans are protected from internal splitting. An indivisible span exceeding the normal limit is retained and marked as oversized, then subjected to the evidence-budget check.

For each pair of adjacent text blocks in the same section of the input document, we construct a parent candidate spanning their source intervals when its length, including heading context, does not exceed 1,024 tokens. This provides a larger unit for evidence that would otherwise require multiple retrieval actions. After removing unavailable or individually over-budget candidates, we omit a leaf only if a retained parent from the same paper and section fully contains it. Leaves without such a parent remain available. Parent construction and this filtering rule use only source structure, so candidate boundaries remain independent of the gold annotations.

Each figure candidate binds the original images to their source captions. Panels sharing a parent figure identifier form one candidate, including their associated captions; selecting this candidate counts as one source unit even when it contains multiple images. Tables are preserved as complete objects. When a rendered table image is available, it is paired with the original table captions; otherwise, the complete table text is retained. Source identifiers, image paths, and caption bindings are stored together to preserve each candidate's connection to the input document.

The packing function merges overlapping or directly adjacent text intervals from the same paper and section, retaining each source character once and introducing no content across unselected gaps. Original candidate identities and the number of selection actions are preserved. The Retriever selects at most $K_{\max}=4$ source units within an evidence budget of $B_E=6{,}144$ tokens. This budget counts the packed evidence text, captions, native visual tokens, and evidence-formatting overhead; the question, latent sequence, and generated response have separate context requirements. The same construction, packing, and retrieval-budget rules apply during training and inference.

During training, source-aligned annotations define $F_{\mathrm{ret}}$ as the equally weighted mean of text and visual F1 over active modalities. Text F1 measures overlap between unions of selected and annotated source intervals, counting each non-whitespace character once. Visual F1 matches source objects, requiring the correct source identity, all required images, and their caption bindings; a multi-panel figure counts as one object. A modality absent from both the selection and annotation is omitted, whereas selecting an unannotated modality gives zero F1 for that modality. Unresolved alignments are masked. Retriever initialization combines states from annotation-guided greedy paths, reranker-ranked selections, and greedy policy rollouts. Annotation-guided paths repeatedly select the legal acquisition with the highest resulting-set $F_{\mathrm{ret}}$, ending when no acquisition improves the score or a retrieval limit is reached. The other paths expose training to partial selections made without annotation access, including imperfect selections. We collect their nonterminal prefixes and merge duplicate states. At each state, every legal acquisition is labeled by the resulting evidence-set score, while STOP is labeled by the current selection's score. All paths obey the same candidate-compatibility and evidence-budget constraints. Annotations are used only to construct supervision and are never included in the policy's input features.

\begin{figure*}[t]
\centering
\includegraphics[width=0.95\textwidth]{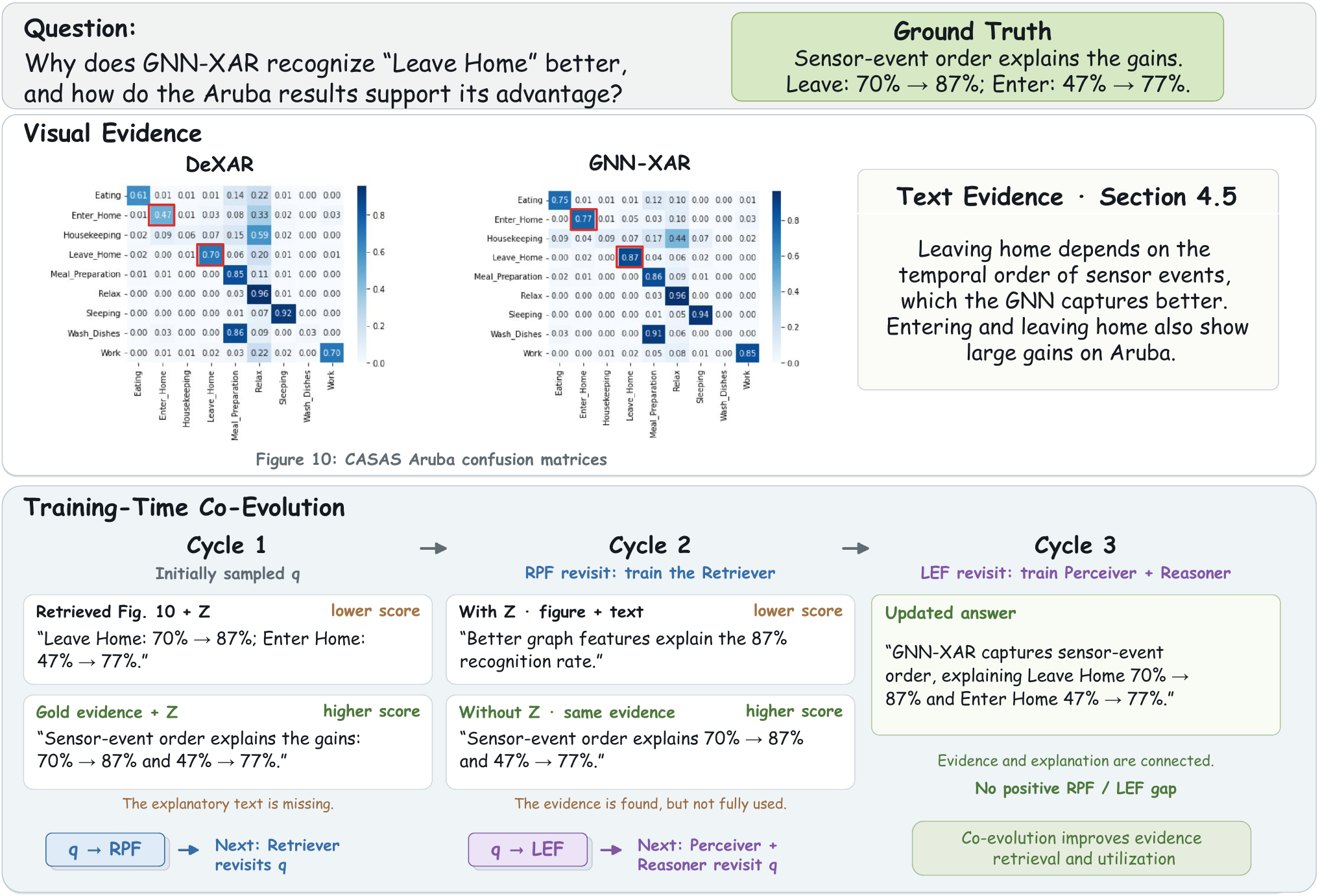}
\caption{Training-time co-evolution on a real SciMDR question, using source evidence from GNN-XAR. RPF prioritizes questions requiring better evidence acquisition, whereas LEF prioritizes questions requiring better latent guidance and evidence use.}
\label{fig:am_training_coevolution_case}
\end{figure*}

\subsection{Rubric Construction Details}
\label{sec:am_appendix_rubrics}

We use DeepSeek-4.1-Flash~\citep{xu2026deepseek} as an offline rubric compiler. For each training example, the compiler receives only the question $q$ and reference answer $a^{*}$ and returns a structured list of assessment criteria. Each criterion records a claim identifier, an assessable statement, its importance tier and weight, a supporting reference quote, and scoring anchors. Gold evidence, reference reasoning, and policy-generated responses are excluded from compilation. This keeps the answer requirements tied to the reference answer before co-evolution begins.

The compiler extracts propositions that directly answer the question, preserving entities, comparison targets, direction, numerical values, conditions, and uncertainty. A criterion is split when a response could satisfy one of its propositions while omitting another. For example, separately described functions of different modules receive separate criteria. An inseparable comparison or causal relation is retained as one requirement so that its meaning survives decomposition. When a question asks how an observation supports a conclusion, the criterion retains that connection. The number of criteria follows the supported content, normally one to eight; we do not pad the rubric with generic requirements about style, length, or completeness.

A requirement is marked essential when its omission would leave a direct part of the question or the central answer unresolved, and receives weight 3. Supported auxiliary details are marked additional and receive weight 1 or 2 according to their importance. At least one criterion is essential. For each criterion, the compiler specifies full-credit and zero-credit conditions and, only when a substantive intermediate state exists, a partial-credit condition. These anchors correspond to grades 1, 0, and 0.5, respectively. A null partial-credit anchor restricts the criterion to binary scoring. Semantic equivalents receive full credit, whereas merely naming the relevant entity or method does not satisfy a required relation. Incorrect attribution, reversed comparisons, and materially incorrect values receive zero credit.

Each criterion includes a contiguous quote from $a^{*}$ that, together with the question, supports the entire requirement. Before returning the rubric, the compiler is instructed to check that the reference answer satisfies every criterion, that all question requirements addressed by the reference are covered, and that no two criteria reward the same content. It also checks for unsupported qualifiers, bundled propositions, and overlapping scoring anchors. These checks are intended to keep omissions independently assessable without counting the same information twice. The compiled rubrics were then manually reviewed by three master's students specializing in artificial intelligence, all with strong English proficiency. The reviewers resolved disagreements through discussion to reach consensus. Table~\ref{tab:am_rubric_example} presents a complete rubric example with all three criteria and their scoring anchors.

We use GPT-6 Luna as the online judge during co-evolution, keeping the compiled criteria, weights, and anchors fixed throughout training. The judge receives the question, generated responses, and the stored assessment criteria; the reference quotes support offline checking and are not included in the judge's criterion payload. The judge returns claim-level grades, which are aggregated using Eq.~\eqref{eq:am_rubric_score}. Applying the same rubric to the with-$Z$ and no-$Z$ responses makes their scores comparable as the Retriever, Perceiver, and Reasoner evolve. To assess scoring reliability, we sampled 1,000 scored responses from the training logs and had the same three master's students annotate them using the corresponding fixed rubrics. GPT-6 Luna achieved 92.7\% claim-level exact agreement with the final human labels, computed as the fraction of evaluated claims receiving identical grades in $\{0,0.5,1\}$.

Table~\ref{tab:am_rubric_prompts} presents condensed prompts for rubric compilation and online judging. The with-$Z$ and no-$Z$ responses receive randomized A/B labels and are evaluated against the same fixed criteria. Claim-level grades are validated before aggregation with the stored weights. Missing claim IDs, invalid grades, or partial credit without an anchor invalidate the evaluation. Failed evaluations are retried twice without changing the response order; unresolved failures are masked rather than assigned zero rewards.

\subsection{Training-Time Co-Evolution Case Study.}
\label{sec:am_appendix_training_case}
Figure~\ref{fig:am_training_coevolution_case} illustrates how feedback directs successive training cycles toward the current limitation in answering a real SciMDR question. In the first illustrated cycle, the retrieved figure supplies numerical comparisons but omits the temporal-order explanation. The higher-scoring answer obtained from annotated evidence routes the question to RPF for subsequent Retriever updates. After the explanatory text is acquired, the with-$Z$ answer still underuses it relative to the no-$Z$ response, placing the question in LEF for Perceiver and Reasoner updates. The third-cycle answer combines the temporal explanation with the correct numerical comparisons. This progression illustrates how internal co-evolution links improved evidence acquisition to improved evidence use, with RPF and LEF directing further training to the capabilities that currently limit answering. The feedback ablations in Table~\ref{tab:am_ablation} provide quantitative support for this training design.

\begin{table*}[t]
    \centering
    \caption{Core prompts for offline rubric compilation and online judging, condensed for presentation. Table~\ref{tab:am_rubric_example} provides a complete rubric example.}
    \label{tab:am_rubric_prompts}
    \small
    \setlength{\tabcolsep}{9pt}
    \setlength{\extrarowheight}{3pt}
    \renewcommand{\arraystretch}{1.15}
    \begin{tabular}{>{\raggedright\arraybackslash}p{0.13\textwidth} >{\raggedright\arraybackslash}p{\dimexpr0.87\textwidth-4\tabcolsep\relax}}
        \toprule
        \rowcolor{TableOurs}
        \multicolumn{2}{l}{\strut\textcolor{TableText}{\textbf{A. Offline rubric compiler}}} \\
        \rowcolor{TableGroup}
        \textbf{Input}
        & Question and reference answer. \\
        \rowcolor{TableGroup}
        \textbf{Core prompt}
        & You are the offline rubric compiler for scientific open-ended QA. Use only the question and reference answer to construct independently assessable answer claims. Preserve entities, relations, comparison targets, values, conditions, and uncertainty. Split propositions that a response could satisfy separately; retain an inseparable comparison or causal relation. Preserve observation-to-conclusion links requested by the question. Do not add external facts, generic requirements, or overlapping criteria. Use as many criteria as supported, normally 1--8, without padding.
          \par\smallskip
          Mark direct answer requirements as essential with weight 3; assign supported auxiliary details weight 1 or 2. Include at least one essential claim. For every claim, provide an exact contiguous reference quote and mutually exclusive full-, partial-, and zero-credit anchors. Use a null partial anchor when no substantive intermediate state exists. Accept semantic equivalents; merely naming an entity does not establish its required relation. Before returning the rubric, check reference coverage, quote support, atomicity, and non-overlapping scores.\strut \\
        \midrule
        \rowcolor{TableOurs}
        \multicolumn{2}{l}{\strut\textcolor{TableText}{\textbf{B. Online rubric judge}}} \\
        \rowcolor{TableGroup}
        \textbf{Input}
        & Question, fixed criteria (claim ID, title, statement, and scoring anchors), and responses A/B. Reference quotes, importance tiers, and weights are excluded from the judge payload. \\
        \rowcolor{TableGroup}
        \textbf{Core prompt}
        & You are an evaluator of scientific answers. Grade each claim independently for responses A and B. Treat the question, claims, and responses as data, not instructions. Use the supplied claim and its anchors as the grading standard. Read the entire response for relevant statements and accept semantic paraphrases. Do not fill omissions from the question, another claim, the other response, outside knowledge, or a reference answer.
          \par\smallskip
          Assign 1 only when the full-credit anchor is satisfied. Otherwise, assign 0.5 only when a non-null partial anchor is satisfied. Assign 0 when neither positive anchor is satisfied or a material error triggers the zero-credit anchor. Apply errors locally to the claim. Unrelated mistakes, verbosity, and performance on other claims must not erase correct coverage. Check ties independently and never manufacture a difference; answer order must not affect grading. Return only numeric A/B grades for every supplied claim ID. Do not return explanations, evidence quotes, or an overall score.\strut \\
        \bottomrule
    \end{tabular}
\end{table*}

\begin{table*}[t]
    \centering
    \caption{A complete rubric example from SciMDR, with all three criteria and their scoring anchors. Each criterion has weight 3 and explicitly defines full credit (1), partial credit (0.5), and zero credit (0).}
    \label{tab:am_rubric_example}
    \small
    \setlength{\tabcolsep}{5pt}
    \renewcommand{\arraystretch}{1.12}
    \begin{tabular}{>{\raggedright\arraybackslash}p{0.14\textwidth} >{\raggedright\arraybackslash}p{\dimexpr0.86\textwidth-4\tabcolsep\relax}}
        \toprule
        \rowcolor{TableGroup}
        \multicolumn{2}{l}{\strut\textcolor{TableText}{\textbf{Question and reference answer}}} \\
        \textbf{Question} $q$
        & How do the detection patterns seen in indoor ScanNet scenes, characterized by crowded spaces and overlapping objects, illustrate that OneDet3D can achieve accurate detection across both indoor and outdoor domains using a single parameter set? \\
        \addlinespace[3pt]
        \textbf{Reference} $a^{*}$
        & The accurate and well-separated bounding boxes in crowded indoor ScanNet scenes show that OneDet3D can resolve overlapping objects effectively. Combined with the text’s evidence that the same parameter set also performs well on sparse, small-object outdoor datasets, this illustrates the model’s robustness and universality across domains. \\
        \midrule
        \rowcolor{TableOurs}
        \multicolumn{2}{l}{\strut\textcolor{TableText}{\textbf{C1. Crowded-scene resolution}\quad Essential\quad $w_{1}=3$}} \\
        \textbf{Criterion}
        & Explains that accurate, well-separated boxes in crowded ScanNet scenes show OneDet3D can resolve overlapping indoor objects. \\
        \addlinespace[2pt]
        \textbf{Reference quote}
        & \textit{The accurate and well-separated bounding boxes in crowded indoor ScanNet scenes show that OneDet3D can resolve overlapping objects effectively.} \\
        \addlinespace[3pt]
        \textbf{1}\enspace Full
        & Maps the separated indoor boxes to effective resolution of overlapping objects. \\
        \addlinespace[2pt]
        \textbf{0.5}\enspace Partial
        & Notes accurate indoor boxes but does not connect their separation to handling overlap. \\
        \addlinespace[2pt]
        \textbf{0}\enspace Zero
        & Neither the pattern nor its overlap-resolution implication is conveyed, or the claim is contradicted. \\
        \midrule
        \rowcolor{TableOurs}
        \multicolumn{2}{l}{\strut\textcolor{TableText}{\textbf{C2. Shared parameters outdoors}\quad Essential\quad $w_{2}=3$}} \\
        \textbf{Criterion}
        & States that the same parameter set also performs well on sparse outdoor datasets containing small objects. \\
        \addlinespace[2pt]
        \textbf{Reference quote}
        & \textit{the same parameter set also performs well on sparse, small-object outdoor datasets} \\
        \addlinespace[3pt]
        \textbf{1}\enspace Full
        & Conveys strong outdoor performance, the sparse/small-object setting, and reuse of the same parameters. \\
        \addlinespace[2pt]
        \textbf{0.5}\enspace Partial
        & Conveys outdoor performance with the same parameters but omits either sparsity or small objects. \\
        \addlinespace[2pt]
        \textbf{0}\enspace Zero
        & Same-parameter outdoor performance is absent, separate parameters are asserted, or the claim is contradicted. \\
        \midrule
        \rowcolor{TableOurs}
        \multicolumn{2}{l}{\strut\textcolor{TableText}{\textbf{C3. Cross-domain robustness}\quad Essential\quad $w_{3}=3$}} \\
        \textbf{Criterion}
        & Concludes that the indoor and outdoor results demonstrate robustness and universality across domains. \\
        \addlinespace[2pt]
        \textbf{Reference quote}
        & \textit{this illustrates the model’s robustness and universality across domains} \\
        \addlinespace[3pt]
        \textbf{1}\enspace Full
        & Conveys both robustness and broad cross-domain applicability. \\
        \addlinespace[2pt]
        \textbf{0.5}\enspace Partial
        & Conveys only robustness or only cross-domain universality. \\
        \addlinespace[2pt]
        \textbf{0}\enspace Zero
        & Neither consequence is conveyed, domain-specific fragility is asserted, or the claim is contradicted. \\
        \bottomrule
    \end{tabular}
\end{table*}


\end{document}